\documentclass{article}

\usepackage{arxiv}

\usepackage[utf8]{inputenc} 
\usepackage[T1]{fontenc}    
\usepackage{hyperref}       
\usepackage{url}            
\usepackage{booktabs}       
\usepackage{amsfonts}       
\usepackage{nicefrac}       
\usepackage{microtype}      
\usepackage{lipsum}
\usepackage{graphicx}
\graphicspath{ {./images/} }

\title{Resource-Efficient Adaptation of Large Language Models for Text Embeddings via Prompt Engineering and Contrastive Fine-tuning}

\author{
Benedikt Roth\textsuperscript{1} \quad
Stephan Rappensperger\textsuperscript{1, 2} \quad
Tianming Qiu\textsuperscript{1, 2} \quad
Hamza Imamović\textsuperscript{1,2} \quad
Julian Wörmann\textsuperscript{1, 2} \quad\\
\textbf{Hao Shen\textsuperscript{1, 2}} \\
\textsuperscript{1}fortiss GmbH, Munich, Germany \\
\textsuperscript{2}School of Computation and Information Technology, Technical University of Munich, Germany \\
\texttt{\{roth, rappensperger, qiu, imamovic, woermann, shen\}@fortiss.org}
}

\begin{document}
\maketitle

\begin{abstract}
Large Language Models (LLMs) have become a cornerstone in Natural Language Processing (NLP), achieving impressive performance in text generation. Their token‑level representations capture rich, human‑aligned semantics. However, pooling these vectors into a text embedding discards crucial information. Nevertheless, many non‑generative downstream tasks, like e.g. clustering, classification, or retrieval, still depend on accurate and controllable sentence‑ or document‑level embeddings.
We explore several adaptation strategies for pre‑trained, decoder‑only LLMs: (i) various aggregation techniques for token embeddings, (ii) task‑specific prompt engineering, and (iii) text‑level augmentation via contrastive fine‑tuning. Combining these components yields competitive performance on the English clustering track of the Massive Text Embedding Benchmark (MTEB). An analysis of the attention map further shows that fine‑tuning shifts focus from prompt tokens to semantically relevant words, indicating more effective compression of meaning into the final hidden state.
Our experiments demonstrate that LLMs can be effectively adapted as text embedding models through a combination of prompt engineering and resource‑efficient contrastive fine‑tuning on synthetically generated positive pairs.

All source code, synthetic training data, and model checkpoints used in this study are publicly available at \href{https://github.com/beneroth13/llm-text-embeddings}{\texttt{github.com/beneroth13/llm-text-embeddings}}.
\end{abstract}

\section{Introduction}

In recent years, Large Language Models (LLMs) have emerged as the state-of-the-art in Natural Language Processing (NLP). Owing to their massive amount of training data and parameters, they excel at a wide range of tasks such as reasoning and text generation.
Although primarily trained for next-token prediction using final hidden states and pooling layers, LLMs have increasingly influenced the field of text embedding generation, despite requiring modifications such as removing task-specific output layers.

Text embeddings are vector representations that capture the core semantics of a text. They serve as input for a variety of downstream tasks, including information retrieval, text clustering, recommendation, topic modeling, and Retrieval-Augmented Generation (RAG), as outlined in \cite{Simple_Techniques}.

Early approaches to text embeddings involved simply averaging over word embeddings obtained from traditional methods such as GloVe \cite{Glove} and Word2Vec \cite{word2vec}. 
A major advancement came with Sentence-BERT (SBERT) \cite{SBERT}, which generates sentence embeddings from BERT using Siamese and triplet network architectures. 

A common approach to improve model performance in an unsupervised setting is contrastive learning, which fine-tunes models to generate more semantically meaningful and discriminative embeddings.
In SimCSE, the authors leveraged dropout to create the positive pairs for contrastive learning, demonstrating strong performance on semantic similarity tasks using BERT \cite{gao2022simcsesimplecontrastivelearning}. PromptBERT \cite{jiang2022promptbertimprovingbertsentence} combines prompting with contrastive learning to enhance BERT \cite{devlin2019bertpretrainingdeepbidirectional}, achieving performance beyond that of SimCSE.

Building on BERT, \cite{Stankevi_ius_2024} summarize and evaluate numerous existing methods for Semantic Textual Similarity (STS), text clustering, and classification. Their study explores information extraction from various layers, experiments with different pooling strategies, and applies post-processing techniques. Another notable method is Diagonal Attention Pooling (DITTO)\cite{chen2023dittosimpleefficientapproach}, which addresses the bias of BERT-based sentence embeddings toward uninformative words. 

Recent advancements in text embedding have been primarily driven by two main fields: prompt engineering and fine-tuning. This paper builds upon both approaches. One example of interest in the evaluation of text embeddings is the Massive Text Embedding Benchmark (MTEB) \cite{muennighoff2023mtebmassivetextembedding}, which collects publicly available datasets and downstream tasks, including clustering, classification, retrieval, STS, and many more. Completing this, a recent survey by \cite{cao2024recentadvancestextembedding} investigates the advancements that have contributed to the best-performing models on MTEB. A key strength of MTEB lies in its large diversity of datasets and task types, which demand general and robust embeddings capable of performing well across varied settings. We are especially interested in improving the clustering performance of English data due to its application to product descriptions. \\

Our contributions can be summarized as follows:

\begin{itemize}
    \item \textbf{A resource-efficient pipeline:} We combine clustering-oriented prompts with LoRA-based contrastive fine-tuning to transform small LLMs into text-embedding models on a single 24 GB GPU.
    
    \item \textbf{Systematic evaluation:} We benchmark state-of-the-art STS prompts, introduce clustering-specific prompts, and compare synthetic data-augmentation strategies with a human-curated Natural Language Inference (NLI) dataset.
    
    \item \textbf{Competitive performance:} The resulting models achieve competitive accuracy on English clustering tasks on MTEB, and the method generalizes well to classification tasks.
    
    \item \textbf{Interpretability:} Attention-map analysis reveals how fine-tuning shifts the model's focus toward semantically relevant tokens.
\end{itemize}

All source code, synthetic training data, and model checkpoints used in this study will be made publicly available.

\section{Related Work}

As already mentioned, LLMs have recently become increasingly popular for text embedding due to their large-scale pretraining. \cite{behnamghader2024llm2veclargelanguagemodels} proposed LLM2Vec, an unsupervised pipeline that creates an embedding model from any decoder-only LLM by enabling bidirectional attention, followed by masked token prediction and unsupervised contrastive learning.
By incorporating supervised learning, they achieved state-of-the-art performance on MTEB for a period. 
First, the authors enhanced the attention mechanism of the LLM by enabling bidirectional attention. They then retrained the model using masked token prediction to adapt it to this new attention scheme. Finally, the authors applied contrastive learning to embeddings obtained through mean-pooling of all attention outputs, fine-tuning the model directly in the embedding space to produce meaningful text representations.
Besides fine-tuning, prompting has also become increasingly popular. \cite{jiang2023scalingsentenceembeddingslarge} use prompts to reshape the embedding space, focusing particularly on encouraging the model to condense a sentence into a single word --- example given, \textit{"This sentence: "\mbox{[X]}" means in one word:“} \cite{jiang2023scalingsentenceembeddingslarge}. 
Building on this work, \cite{Simple_Techniques} also leverage LLMs and prompt engineering, testing new prompts and showing that, in a zero-shot inference setting, enforcing an explicit one-word limitation (EOL) improves performance, and the prompt choice significantly affects results. When using efficient fine-tuning via contrastive learning on a human-annotated NLI dataset, they show that the prompt has little impact on STS tasks.
They further introduce new zero-shot prompts, such as \textit{After thinking step by step, this sentence: "\mbox{[X]}" means in one word:“} \cite{Simple_Techniques}, which they refer to \emph{Pretended Chain of Thought}. They also present a second prompt called \emph{Knowledge Enhancement}.
Qwen3 Embedding has been recently released, comprising a family embedding and reranking models \cite{zhang2025qwen3embeddingadvancingtext}. The models explicitly use an EOS token, leveraging its final hidden state as the text embedding. In their technical report, the authors describe the training pipeline as follows: (1) weakly supervised pretraining using large-scale synthetic paired data, (2) supervised fine-tuning with high-quality synthetic and labeled data, and (3) model merging using sampled checkpoints from the previous step \cite{zhang2025qwen3embeddingadvancingtext}. The synthetic data was generated using the Qwen3-32B model, resulting in 150 million pairs for multi-task weakly supervised learning. Filtering the synthetic data by cosine similarity, they reduced it to 12 million high-quality supervised training data pairs. Their methods were evaluated on MTEB \cite{muennighoff2023mtebmassivetextembedding}, achieving state-of-the-art performance.

\section{Methods}

\begin{figure*}[!t]
  \centering
  \includegraphics[width=1.0\textwidth]{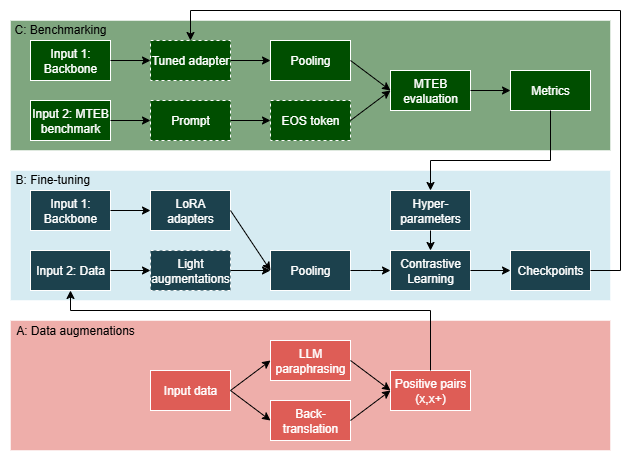}
  \caption{Overview of the proposed pipeline. It comprises three stages: (A) data augmentation via back-translation and LLM paraphrasing, (B) LoRA fine-tuning with contrastive learning, and (C) benchmarking on MTEB with optional prompts, tuned adapters, and EOS tokens. Dashed boxes mark optional components.}
  \label{fig:pipeline}
\end{figure*}

Our methodology follows a three-stage workflow visualized in Figure~\ref{fig:pipeline}. First, we conduct a zero-shot evaluation of token aggregation strategies over pretrained LLM representations, using the English clustering subset of  MTEB. Second, we shape the embedding space through prompting by wrapping clustering-oriented prompts around each input, encouraging representations that better capture text-level semantics. Third, we apply LoRA-based contrastive fine-tuning by adding low-rank adapters to the frozen backbone and optimizing them with a contrastive loss \cite{lora}. During this stage, we compare several positive-pair generation strategies to better align the embeddings produced under the selected prompt.

\subsection{Embedding Extraction \& Pooling}

Given an input sequence, we extract the hidden states from the final transformer layer and derive a sentence vector using one of three token aggregation strategies:
\begin{itemize}
\item \textbf{Mean Pooling} --- Average all token hidden states.
\item \textbf{Last-Token Pooling} --- Use the hidden states of the final token in the sequence.
\item \textbf{EOS-Token Pooling} --- Append an explicit end-of-sequence (EOS) token and use its hidden state.
\end{itemize}

In zero-shot evaluation, EOS-token pooling consistently underperformed compared to the implicit last-token variant. As a result, all subsequent analyses focus on mean and last-token pooling. Mean pooling aggregates information across all positions, while last-token pooling leverages the auto-regressive objective, which encourages the final hidden state to encode the entire context. Although prior work reports promising performance using explicit EOS pooling \cite{Simple_Techniques}, our experiments indicate that this benefit does not generalize to the models and clustering tasks experimented in this study.

\subsection{Prompt-Based Shaping of the Embedding Space}

Prompt engineering is a proven technique for steering LLM representations in STS tasks \cite{jiang2023scalingsentenceembeddingslarge, Simple_Techniques}. To adapt this idea to clustering, we re-evaluated the STS prompts and introduced cluster-specific variants. Table~\ref{tab:prompts_clustering} lists the six prompts considered.

\begin{table}[ht]
  \centering
  {\normalsize
  \begin{tabular}{@{}l@{}}
    \hline
    \textbf{Explicit One-Word Limitation (EOL)} \\
    This sentence: "\mbox{[X]}" means in one word. \\
    \hline
    \textbf{Pretended Chain of Thought (PCoT)} \\
    After thinking step by step, this sentence: "\mbox{[X]}" means in one word. \\
    \hline
    \textbf{Summarize (SUM)} \\
    This sentence: "\mbox{[X]}" can be summarized as. \\
    \hline
    \textbf{Cluster Command Wrapped (CCW)} \\
    This sentence: "\mbox{[X]}" belongs to the following cluster. \\
    \hline
    \textbf{Cluster Command Prepended (CCP)} \\
    Cluster the text: "\mbox{[X]}". \\
    \hline
    \textbf{Question} \\
    Which cluster would you assign the sentence: "\mbox{[X]}" to? \\
    \hline
  \end{tabular}
  }
  \vspace{0.5em}
  \caption{Prompts used in the experimental evaluation. EOL, PCoT, and SUM are adopted from prior work \cite{Simple_Techniques}, while CCW, CCP, and Question are novel contributions specifically designed for clustering tasks. The abbreviated names are used throughout the paper for brevity.}
  \label{tab:prompts_clustering}
\end{table}

We first replicate the three STS prompts --- EOL, Pretended Chain of Thought (PCoT), and Summarize (SUM) --- as introduced by \cite{Simple_Techniques}. We then design two clustering-oriented prompts, Cluster Command Wrapped (CCW), Cluster Command Prepended (CCP), and a question-style prompt (Question). CCP follows the instruction format used in Qwen3-Embedding \cite{zhang2025qwen3embeddingadvancingtext}, which takes the form \texttt{\{Instruction\} \{Query\} <|endoftext|>}.

This format places the instruction before the input, allowing the final token to remain content-dependent. In contrast, CCW wraps the instruction around the sentence, fixing the final token (":") and allowing the attention mechanism to propagate context. We omit an explicit \texttt{<eos>} token, as earlier pooling experiments showed no benefit. By comparing these alternatives, we assess how instruction format and token position influence the quality of the resulting clustering embeddings.

\subsection{Contrastive Fine-Tuning with Augmented Data}

The third stage applies parameter-efficient LoRA fine-tuning to (i) adapt the model to the selected prompt and (ii) optimize it to produce the sentence embedding directly, rather than merely generating an intermediate representation for next-token prediction. Following \cite{Simple_Techniques}, we attach LoRA adapters and train them using a contrastive loss in a self-supervised setting. 
As an initial source of positive pairs, we adopt the wiki1m\_for\_simcse dataset, a one-million-sentence Wikipedia corpus dataset used by SimCSE \cite{gao2022simcsesimplecontrastivelearning}. The motivation for using this dataset follows \cite{behnamghader2024llm2veclargelanguagemodels}, which argues that since LLMs have already been trained on Wikipedia data, they will not acquire new domain knowledge, but only learn the fine-tuned technique. Negative pairs are generated by treating all non-positive samples in each training batch as negatives \cite{oord2019representationlearningcontrastivepredictive}.
We then extended our study by applying a suite of natural-language data augmentation techniques \cite{Bayer_2022,chai2025textdataaugmentationlarge,wang2025comprehensivesurveydataaugmentation}, which are described in the remainder of the section.

\begin{itemize}
\item \textbf{Translation:} We translate each sentence from English into German and then back into English.
\item \textbf{Generative AI:} We prompt an LLM to generate positive text pairs.
\item \textbf{Random Deletion:} We randomly delete each word with a probability of 10\%.
\item \textbf{Random Swap:} We randomly swap two words, performing this operation only once per text.
\item \textbf{Character-Level Noise:} We simulate typographical errors by randomly swapping two adjacent characters within a word, with a probability of 5\%.
\end{itemize}

We progressively diversified the source of positive pairs as follows.
First, we performed back-translation (English $\rightarrow$ German $\rightarrow$ English) on the Wikipedia corpus and fine-tuned a model on the resulting positive pairs. We then introduced lexical perturbations to increase surface-level variation. Next, we fine-tuned a model on a human-annotated NLI corpus previously used by \cite{Simple_Techniques} and created by \cite{gao2021nli_simcse,bowman2015largeannotatedcorpuslearning, A_Broad-Coverage_Challenge_Corpus}, which provides high-quality positive pairs. Finally, synthetic positive pairs were generated through LLM-based paraphrasing, using the prompts listed in Table~\ref{tab:prompts_augmenting}, and used them in a separate fine-tuning stage.

\begin{table}[ht]
  \centering
  {\normalsize
  \begin{tabular}{@{}l@{}}
    \hline
    Rewrite the following sentence with more detail, keeping the original meaning.\\
    Respond with only the rewritten sentence: "\mbox{[X]}" \\
    \hline
    Paraphrase this sentence using simpler language.\\
    Respond with only the rewritten sentence: "\mbox{[X]}" \\
    \hline
    You’re a high school teacher. Explain this sentence to your students in your own words.\\
    Respond with only the explanation: "\mbox{[X]}" \\
    \hline
    Write a metaphor that expresses the same idea as this sentence.\\
    Respond with only the metaphor: "\mbox{[X]}" \\
    \hline
    After reading this sentence, what is a natural question someone might ask?\\
    Respond with only the question: "\mbox{[X]}" \\
    \hline
  \end{tabular}
  }
  \vspace{0.5em}
  \caption{Prompt templates used to generate synthetic positive pairs from sentences in the Wikipedia corpus. Each prompt elicits a paraphrase or semantically equivalent variant, enriching contrastive training with linguistic diversity.}
  \label{tab:prompts_augmenting}
\end{table}

\section{Results}

In this section, we address three questions:
(i) Which pooling strategy produces the strongest zero-shot clustering performance?
(ii) To what extent do clustering-oriented prompts improve these baselines?
(iii) Which combination of fine-tuning, dataset, and augmentation strategy provides the best results?
Unless stated otherwise, scores are averaged over the 21 English clustering datasets in MTEB v1.38.30. We also provide classification results to assess the generality of our findings. In line with the established MTEB protocol, we report single-run scores without confidence intervals or variance estimates to ensure direct comparability with prior work. Because MTEB fixes dataset splits, random seeds, and evaluation scripts, repeated evaluations of the same model yield identical results. Experiments were conducted on a single NVIDIA RTX A5000 (24 GB), except for models exceeding one billion parameters, which were tested using multiple GPUs. Our study focuses on two smaller LLMs, namely Llama-3.2-1B \cite{grattafiori2024llama3herdmodels} and Qwen3-0.6B \cite{yang2025qwen3technicalreport}.

\subsection{Embedding Extraction \& Pooling}

The zero-shot baseline was obtained by comparing three token aggregation strategies across Llama-3.2-1B and Qwen3-0.6B. As shown in Table~\ref{tab:zero-shot-clustering}, mean pooling delivers the highest average clustering accuracy, outperforming last-token pooling by approximately 12\,\% and EOS-token pooling by around 20\,\% across both models. The Llama model also shows a consistent advantage over Qwen, reflecting its larger parameter count.

\begin{table}[ht]
  \centering
  {\normalsize
  \begin{tabular}{@{}lccc@{}}
    \hline
    \textbf{Model} & \textbf{Mean‑pooling} & \textbf{Last‑token} & \textbf{EOS‑token} \\
    \hline
    Llama‑3.2‑1B & \textbf{45.8\,\%} & 33.2\,\% & 23.4\,\% \\
    Qwen3‑0.6B   & \textbf{39.8\,\%} & 28.9\,\% & 22.8\,\% \\
    \hline
  \end{tabular}
  }
  \vspace{0.5em}
  \caption{Zero-shot clustering accuracy on 21 English datasets from MTEB. Mean pooling consistently outperforms last-token and EOS-token aggregation, across both models.}
  \label{tab:zero-shot-clustering}
\end{table}

\subsection{Prompt-Based Shaping of the Embedding Space}

Prompting alters the embedding space unevenly across different pooling strategies, as shown in Table~\ref{tab:zero-shot-clustering-with-prompts}. Mean pooling yields marginal or even detrimental effects. Intuitively, that is likely because many tokens remain constant across prompts, pulling the mean embedding toward shared content and outweighing improvements in the original text's token representation. By contrast, last-token pooling benefits substantially. This behavior likely stems from the model's tendency to prepare an answer when prompted, concentrating relevant information in the final token. The fact that semantically equivalent prompts yield significantly different results highlights the limited robustness of current LLMs. 

\begin{table}[t]
  \centering
  \begin{tabular}{lcc}
    \hline
    \textbf{Model} & \textbf{Mean} & \textbf{Last} \\
    \hline
    Llama-3.2-1B              & 45.8\,\% & 33.2\,\% \\
    Llama-3.2-1B + EOL        & \textbf{46.1\,\%} & 43.2\,\% \\
    Llama-3.2-1B + PCoT       & 44.6\,\% & 42.8\,\% \\
    Llama-3.2-1B + SUM        & 43.9\,\% & 36.7\,\% \\
    Llama-3.2-1B + CCW        & 45.3\,\% & \textbf{46.8\,\%} \\
    Llama-3.2-1B + CCP        & 43.5\,\% & 32.8\,\% \\
    Llama-3.2-1B + Question   & 43.4\,\% & 30.8\,\% \\
    \hline
    Qwen3-0.6B                & \textbf{39.8\,\%} & 28.9\,\% \\
    Qwen3-0.6B + EOL          & 38.3\,\% & 26.9\,\% \\
    Qwen3-0.6B + PCoT         & 38.2\,\% & 24.3\,\% \\
    Qwen3-0.6B + SUM          & 38.5\,\% & 28.0\,\% \\
    Qwen3-0.6B + CCW          & 38.6\,\% & \textbf{30.7\,\%} \\
    Qwen3-0.6B + CCP          & 38.9\,\% & 29.8\,\% \\
    Qwen3-0.6B + Question     & 37.5\,\% & 24.8\,\% \\
    \hline
  \end{tabular}
  \caption{Clustering accuracy with and without prompt-based input formatting. Results are reported using mean and last-token pooling across prompts. The CCW prompt leads to the largest gain when combined with last-token pooling.}
  \label{tab:zero-shot-clustering-with-prompts}
\end{table}

On Llama-3.2-1B, the CCW prompt increases last-token accuracy from 33.2\,\% to 46.8\,\%, a gain of 13.6 percentage points with barely any additional inference cost. Other instruction-style prompts (EOL, PCoT) also improve performance, while the interrogative variant (Question) degrades it --- suggesting that commands produce more consistent representations than questions. Qwen3-0.6B exhibits similar but smaller gains, as none of the tested prompts closes the gap to Llama --- likely due to Qwen's lower capacity. We were unable to identify an equally effective prompt for Qwen. The results further suggest that wrapping the prompt around the input may be beneficial.

\subsection{Contrastive Fine-Tuning with Augmented Data}

\begin{table*}[th]
  \centering
  \small
  \begin{tabular*}{\textwidth}{@{}l @{\extracolsep{\fill}} c c c c@{}}
    \hline
    \textbf{Method}
      & \parbox[c]{2cm}{\centering\textbf{Mean‑pooling}\\\textbf{Llama‑3.2‑1B}}
      & \parbox[c]{2cm}{\centering\textbf{Last‑token}\\\textbf{Llama‑3.2‑1B}}
      & \parbox[c]{2cm}{\centering\textbf{Mean‑pooling}\\\textbf{Qwen3‑0.6B}}
      & \parbox[c]{2cm}{\centering\textbf{Last‑token}\\\textbf{Qwen3‑0.6B}} \\
    \hline
    Base                   
      & 45.8\,\% & 33.2\,\% & 39.8\,\% & 28.9\,\% \\
    EOL                 
      & 46.1\,\% & 43.2\,\% & 38.3\,\% & 26.9\,\% \\
    CCW                 
      & 45.3\,\% & 46.8\,\% & 38.6\,\% & 30.7\,\% \\
    CCP                 
      & 43.5\,\% & 32.8\,\% & 38.9\,\% & 29.8\,\% \\
    Base + nli    
      & 49.2\,\% & 47.8\,\% & 46.8\,\% & 47.4\,\% \\
    CCW + nli   
      & 52.5\,\% & 52.2\,\% & \textbf{49.2\,\%} & 49.6\,\% \\
    CCP + nli    
      & 49.5\,\% & 49.7\,\% & 46.2\,\% & 48.9\,\% \\
    CCW + wiki‑translate
      & 51.2\,\% & 52.2\,\% & 47.1\,\% & 49.4\,\% \\
    CCP + wiki‑translate
      & 47.0\,\% & 45.3\,\% & 44.4\,\% & 43.8\,\% \\
    CCW + wiki‑translate + aug
      & 51.7\,\% & 52.6\,\% & 46.8\,\% & 47.7\,\% \\
    CCP + wiki‑translate + aug
      & 47.2\,\% & 46.5\,\% & 43.4\,\% & 45.6\,\% \\
    CCW + wiki‑prompt   
      & \textbf{52.7\,\%} & \textbf{54.5\,\%} & 48.8\,\% & \textbf{50.9\,\%} \\
    CCP + wiki‑prompt   
      & 48.4\,\% & 51.8\,\% & 45.3\,\% & 49.0\,\% \\
    \hline
  \end{tabular*}
  \caption{Comparison of zero-shot, prompt-based, and fine-tuned clustering performance across various training configurations. Each row corresponds to a specific combination of prompt template and training dataset. The best result for Llama-3.2-1B is obtained using CCW prompt with LLM-generated positive pairs from the wiki-prompt dataset.}
  \label{tab:method-comparison-fine-tuning}
\end{table*}

For fine-tuning, we constrained the search space to two clustering prompts (CCW and CCP), two pooling strategies (mean and last-token), and the data configuration introduced in the Methods section. Results are summarized in Table~\ref{tab:method-comparison-fine-tuning}.

Consistent with prior findings, prompts benefit the last-token variant substantially more than mean pooling.

Training LoRA adapters on the human-curated NLI corpus without a prompt improves performance for both models, yet it does not surpass prompt-based fine-tuning. On Llama, the gap is particularly pronounced --- fine-tuning NLI without a prompt achieves 47.8\,\%, barely outperforming CCW prompting alone at 46.8\,\%.

Using Wikipedia sentences generated through back-translation, referred to as "wiki-translate", yields lower performance than NLI, and the gap becomes even larger when combined with CCP. Adding lexical and character-level noise as data augmentation leads to modest improvements for Llama but often degrades performance for Qwen.

The best results are obtained using LLM-generated positive pairs, referred to as “wiki-prompt”. When combined with CCW, last-token accuracy reaches 54.5\,\% for Llama and 50.9\,\% for Qwen, representing the overall best scores. Mean pooling also performs strongly, peaking at 52.7\,\% for Llama on the same synthetic dataset. For Qwen, mean pooling achieves its best result on NLI data.

CCW, which wraps the sentence, consistently outperforms CCP across all datasets. Synthetic positive pairs generated by an LLM rival human-annotated NLI samples, while remaining domain-agnostic and highly scalable.

Detailed results for each task, including v\_measure scores and standard deviations, are reported in Appendix \ref{sec:appendix_detailed} for the CCW + wiki-prompt setting using Llama with last-token pooling. Also, all hyperparameter details are provided in Appendix \ref{sec:appendix_hyperparams}.

\subsection{Comparison to other models on MTEB(eng, v2) Clustering}

Table~\ref{tab:clustering-comparison} compares our best model against a wide set of available embedding models. Proprietary systems such as \textit{gemini-embedding-001}~\cite{lee2025geminiembeddinggeneralizableembeddings} (59.39\,\%) and large-scale open models like \textit{gte-Qwen2-7B-instruct}~\cite{alibaba2024gteqwen2} (58.97\,\%) and \textit{Qwen3-Embedding-8B} (58.57\,\%) currently define the state of the art. Slightly smaller variants such as \textit{Qwen3-Embedding-4B} (57.51\,\%) and \textit{GritLM-7B}~\cite{muennighoff2025generativerepresentationalinstructiontuning} (50.82\,\%) further illustrate the performance advantages of scaling dedicated embedding models.  

Within this landscape, our 1.3B-parameter checkpoint achieves 50.97\,\%, outperforming general-purpose models of comparable or smaller size such as \textit{GIST-large-Embedding-v0}~\cite{solatorio2024gistembedguidedinsampleselection} (48.84\,\%), \textit{multilingual-e5-large-instruct}~\cite{wang2024multilinguale5textembeddings} (49.89\,\%), and \textit{NV-Embed-v2}~\cite{lee2025nvembedimprovedtechniquestraining} (47.66\,\%). Compared to the Qwen3-Embedding family, our approach trails the 4B and 8B variants, but remains close to \textit{Qwen3-Embedding-0.6B}, despite being trained with far less compute and without task-specific prompting.

Importantly, this performance is achieved with a lightweight LoRA adaptation on Llama-3.2-1B, requiring under one GPU-hour on a single RTX A5000. This demonstrates that competitive clustering quality can be obtained without scaling to multi-billion-parameter embedding-specialized models, reinforcing the efficiency and adaptability of our method. 

\begin{table}[h]
  \centering
  \begin{tabular}{lcc}
    \hline
    \textbf{Model} & \textbf{Param.} & \textbf{Clustering} \\
    \hline
    NV-Embed-v2                & 7.8B  & 47.66\,\% \\
    GIST-large-Embedding-v0    & 0.3B  & 48.84\,\% \\
    multilingual-e5-large-in.  & 0.6B  & 49.89\,\% \\
    GritLM-7B                  & 7.2B  & 50.82\,\% \\
    gte-Qwen2-1.5B-instruct    & 1.5B  & 53.54\,\% \\
    gte-Qwen2-7B-instruct      & 7.6B  & 58.97\,\% \\
    Qwen3-Embedding-0.6B       & 0.6B  & 54.05\,\% \\
    Qwen3-Embedding-4B         & 4B    & 57.51\,\% \\
    Qwen3-Embedding-8B         & 8B    & 58.57\,\% \\
    gemini-embedding-001       & –     & 59.39\,\% \\
    Our best (Llama-3.2-1B + CCW) & 1.3B  & 50.97\,\% \\
    \hline
  \end{tabular}
  \caption{Comparison of clustering performance between our best Llama model, the Qwen3‑Embedding family, and other competitors. Results are reported for MTEB(eng, v2) Clustering, introduced by \cite{enevoldsen2025mmtebmassivemultilingualtext}.}
  \label{tab:clustering-comparison}
\end{table}

\begin{table}[ht]
  \centering
  \begin{tabular}{lcc}
    \hline
    \textbf{Model} & \textbf{Mean} & \textbf{Last} \\
    \hline
    Llama-3.2-1B                  & 72.5\,\% & 68.1\,\% \\
    Llama-3.2-1B + Prompt         & 73.0\,\% & 70.3\,\% \\
    Llama-3.2-1B + Prompt + FT    & 74.2\,\% & 75.4\,\% \\
    Qwen3-0.6B                    & 70.2\,\% & 63.0\,\% \\
    Qwen3-0.6B + Prompt           & 70.3\,\% & 64.1\,\% \\
    Qwen3-0.6B + Prompt + FT      & 72.4\,\% & 72.5\,\% \\
    \hline
  \end{tabular}
  \caption{Classification accuracy on the English subset of MTEB using the same configuration as in the clustering experiment. No hyperparameter adjustment was done. A classification-style prompt structurally aligned with the best-performing clustering prompt (CCW) was used.}
  \label{tab:classification-overview}
\end{table}

\subsection{Classification as a Downstream Task}

To evaluate the generality of our pipeline, we applied it to the English classification subset of MTEB without modifying hyperparameters or prompt templates. Guided by insights from clustering experiments, we used the prompt: \textit{"This sentence: "\mbox{[X]}" can be classified as:"}. Datasets with titles containing legal were excluded due to licensing constraints, and YahooAnswersTopicsClassification was omitted because empty entries triggered input formatting errors with the Qwen tokenizer. Table~\ref{tab:classification-overview} reports the macro-average accuracy across the remaining tasks.

The observed performance pattern closely mirrors that of the clustering tasks. In the zero-shot setting, mean pooling consistently outperforms last-token pooling, and Llama-3.2-1B maintains a slight advantage over Qwen3-0.6B. Introducing the CCW-style prompt yields modest improvements, again more pronounced for last-token embeddings. Fine-tuning with LoRA using the same hyperparameters as in the clustering experiments increases last-token accuracy to 75.4\,\% for Llama and 72.5\,\% for Qwen, establishing this as the best overall configuration. Notably, these improvements are obtained without classification-specific prompt tuning or hyperparameter optimization, highlighting the transferability of the pipeline across task types.

We further evaluate a key advantage of our method --- its ability to incorporate domain-specific knowledge through targeted fine-tuning. To this end, we fine-tune Llama-3.2-1B on two MTEB classification datasets with labeled training splits: EmotionClassification and ToxicConversationsClassification\cite{muennighoff2023mtebmassivetextembedding}.
Based on prior results showing its superiority in fine-tuned settings, we use last-token pooling exclusively.
For each task, we construct a domain-specific contrastive training set by prompting an LLM to generate paraphrases for each sentence in the task’s training set, thereby creating positive pairs. Fine-tuning on these smaller classification datasets is even faster than in the clustering setting. As before, less than a full epoch is required, with performance typically peaking after approximately 100 optimization steps.

\begin{table}[h]
  \centering
  \begin{tabular}{lcc}
    \hline
    \textbf{Method} & \textbf{Emotion} & \textbf{Toxic} \\
    \hline
    Last                        & 29.26\,\% & 63.53\,\% \\
    Last + Prompt               & 42.81\,\% & 63.19\,\% \\
    Fine-tuned (general)        & 45.33\,\% & 67.38\,\% \\
    Fine-tuned (task)           & \textbf{49.78\,\%} & \textbf{68.93\,\%} \\
    \hline
  \end{tabular}
  \caption{Classification accuracy on EmotionClassification and ToxicConversationsClassification. “Last” denotes the last-token embedding without a prompt; “+ prompt” adds a classification-style prompt. “Fine-tuned (general)” refers to LoRA on the wiki-prompt corpus; “Fine-tuned (task)” uses LLM-generated positive pairs from each task’s training set.}
  \label{tab:emotion_toxic_comparison}
\end{table}

\begin{figure*}[ht]
  \centering
  \includegraphics[width=1.0\textwidth]{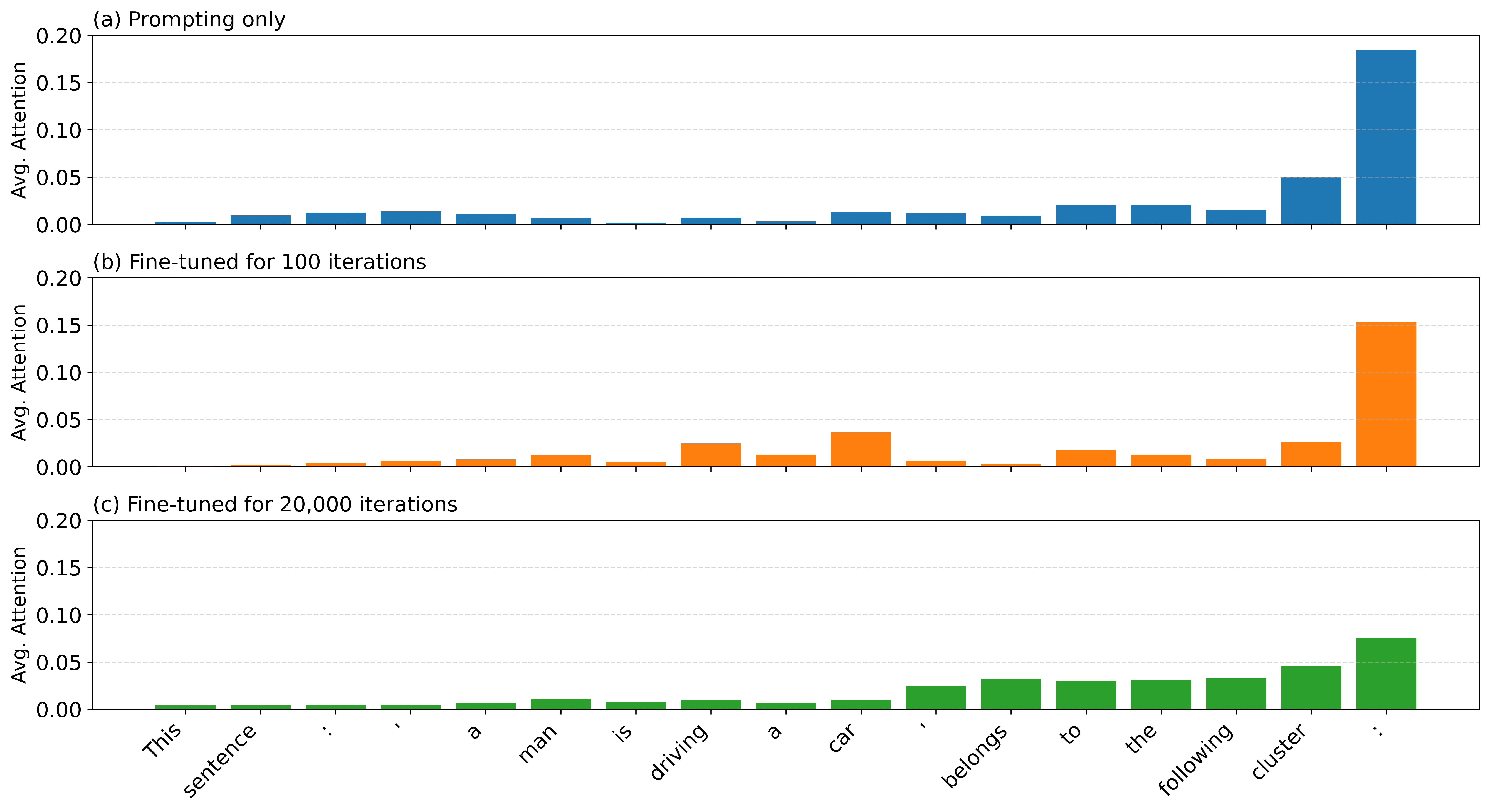}
  \caption{Average attention received by each input token from the final token in the last transformer layer of Llama-3.2-1B, prompted with \textit{"This sentence: ‘a man is driving a car’ belongs to the following cluster:"}. Bars indicate values for at three stages of training: prompt-only, after 100 fine-tuning steps, and after 20,000 steps. Fine-tuning initially shifts attention to semantically meaningful tokens (“man”, “driving”, “car”), while prolonged training leads to a more uniform distribution.}
  \label{fig:attention_plot}
\end{figure*}

Table~\ref{tab:emotion_toxic_comparison} illustrates a consistent pattern. The classification prompt leads to a substantial improvement on the Emotion task but slightly deteriorates for Toxicity. Fine-tuning on the general wiki-prompt corpus enhances performance across both tasks. However, fine-tuning on the task-specific data achieves the highest scores, improving Emotion by 4.5\,\% and Toxicity by 1.6\,\% over the general model. While the extent of improvement varies, the addition of task-relevant supervision consistently results in further gains.

\subsection{Analysis of Attention}

To investigate how LoRA fine-tuning alters the model's internal focus, we follow a method similar to that of \cite{Simple_Techniques}. Attention maps were generated for the probe sentence \textit{"a man is driving a car"} and validated using additional sentences sampled from MTEB. Additional examples and detailed discussion are provided in Appendix \ref{sec:attention}. The analysis examines the outgoing attention distribution of the final token in Llama-3.2-1B with the CCW prompt, comparing three settings illustrated in Figure~\ref{fig:attention_plot}: panel (a) shows the zero-shot model, panel (b) presents the best checkpoint after approximately 100 optimization steps, and panel (c) displays an over-trained checkpoint after 20,000 steps.

In all cases, the final token directs most of its attention to itself, a well-known behavior in causal transformers that preserves the accumulated context through residual connections. The key difference lies in the distribution of the remaining attention. After moderate fine-tuning, attention shifts away from the prompt tokens toward the content words man, \emph{driving}, and \emph{car}, emphasizing the sentence’s subject, predicate, and object --- the elements most informative for downstream semantic tasks. With extended training, the focused attention dissipates; attention becomes more uniformly distributed, and model performance declines accordingly. These observations suggest that an appropriately timed fine-tuning phase encourages the model to encode task-relevant semantics into the final hidden state, while over-training diffuses this information and diminishes embedding quality.

\section{Conclusion}

This work focuses on turning pretrained LLMs into strong sentence-embedding models.
We show that parameter-efficient fine-tuning, such as LoRA steered by a well-chosen prompt, can achieve high-quality embeddings.
Additionally, sentence-level data augmentation like synthetic positive pairs enables promising contrastive self-supervised fine-tuning without manually labeled data.
We raised Llama-3.2-1B and Qwen3-0.6B to highly competitive performance on the English clustering subset of MTEB, rivaling a dedicated embedding baseline in under an hour on a single RTX A5000.
Attention maps indicate a more interpretable adaptation to downstream tasks.

\section*{Limitations}

The current framework is limited to task-specific prompt formats like clustering or classification, English-only text data, and full-precision models. It does not yet incorporate task-agnostic prompt design, multilingual or cross-modal applications, or quantization methods for reducing resource consumption even further. 
Furthermore, the attention analysis presented here should not be interpreted as providing causal explanations; instead, it offers descriptive evidence. While the analysis cannot establish definitive cause-and-effect relationships, the consistent patterns observed across a broad range of test sentences increase confidence that the model tends to focus on the input to be encoded, with particular emphasis on the most relevant tokens.

\section*{Acknowledgements}
This work has been supported by the Bavarian Ministry of Economic Affairs, Regional Development and Energy (STMWi) through the KIMaKu project DIK-2308-0036//DIK0547/01.

We used language and writing support tools (ChatGPT, DeepL Write, and Grammarly) to improve readability and grammar. ChatGPT was also employed to assist with code generation and debugging. The authors remain solely responsible for the content of this work.

\bibliographystyle{unsrt}  
\bibliography{references} 

\appendix

\clearpage

\section{Hyperparameter-Tuning}
\label{sec:appendix_hyperparams}

Table~\ref{tab:hyperparameter} lists the hyperparameters explored during fine-tuning. The final settings used in our experiments are highlighted in bold. We initially trained LoRA adapters on two attention projections, namely the query and value projections. However, we observed higher scores after extending optimization to include six projections in total. These additional projections also influence the feed-forward pathway and include the output, gate, up, and down projections. This broader parameter coverage is particularly well-suited for embedding tasks, as it allows the model to adapt both the attention mechanism and token-mixing components.

\begin{table}[h]
  \centering
  \small
  \begin{tabular}{ll}
    \hline
    \textbf{Hyperparameter} & \textbf{Tested values} \\
    \hline
    Dropout            & \textbf{0.05}, 0.1, 0.2, 0.5 \\
    Learning rate      & \(1\times10^{-5}\), \(\mathbf{5\times10^{-5}}\), \(5\times10^{-4}\) \\
    Batch size         & 20, 60, \textbf{120}, 500 \\
    Temperature        & 1, 0.5, \textbf{0.2}, 0.1, 0.05 \\
    Projection layers  & 2, \textbf{6} \\
    \hline
  \end{tabular}
  \caption{Hyperparameter search space used for LoRA fine-tuning. The tested values for each parameter are listed, with the final configuration highlighted. For projection layers, both two-layer and six-layer settings were evaluated.}
  \label{tab:hyperparameter}
\end{table}

All fine-tuning experiments were conducted on a single NVIDIA RTX A5000 GPU with 24 GB of memory. Peak validation performance typically emerged within 0.1 to 0.5 epochs, corresponding to approximately 100 optimization steps or 12,000 examples at the selected batch size of 120. In some cases, performance plateaued later, between 500 and 2,500 steps. To address this, we saved intermediate checkpoints and selected the best-performing model post hoc. In practice, fine-tuning runs completed under 30 minutes, demonstrating the method's efficiency and suitability for low-resource environments. All additional hyperparameters can be found in the code in their final ideal state.

\section{Results}

This appendix section provides supplementary experimental results and analyses that complement the findings in the main text. We report additional zero-shot comparisons across model sizes, detailed per-dataset performance with standard deviations, and further qualitative analyses such as attention visualizations.

\subsection{Zero-Shot Comparison by Model Size}

In a strictly zero-shot setting, we also evaluated larger checkpoints: Llama-3.2-3B, Qwen3-4B, and Qwen3-8B. Only the CCW prompt was used in this comparison, as shown in Table~\ref{tab:tests-versus-larger-models-zero-shot}.

\begin{table}[h]
  \centering
  \begin{tabular}{lcc}
    \hline
    \textbf{Model} & \textbf{Mean} & \textbf{Last} \\
    \hline
    Llama-3.2-1B              & 45.8\,\% & 33.2\,\% \\
    Llama-3.2-1B + Prompt     & 45.3\,\% & 46.8\,\% \\
    Llama-3.2-3B              & 45.4\,\% & 33.4\,\% \\
    Llama-3.2-3B + Prompt     & 45.5\,\% & 46.8\,\% \\
    Qwen3-0.6B                & 39.8\,\% & 28.9\,\% \\
    Qwen3-0.6B + Prompt       & 38.6\,\% & 30.7\,\% \\
    Qwen3-4B                  & 40.8\,\% & 28.3\,\% \\
    Qwen3-4B + Prompt         & 40.3\,\% & 46.3\,\% \\
    Qwen3-8B                  & 35.3\,\% & 26.7\,\% \\
    Qwen3-8B + Prompt         & 35.4\,\% & 44.1\,\% \\
    \hline
  \end{tabular}
  \caption{Zero-shot clustering score across different model sizes, with and without the CCW prompt. The results highlight how model scaling and prompt-based input formatting influence performance, both mean and last-token pooling. The CCW prompt is consistently applied.}
  \label{tab:tests-versus-larger-models-zero-shot}
\end{table}

Scaling Llama from 1 billion to 3 billion parameters yields virtually no change in either pooling metric, suggesting that the Llama-3.2-1B model already saturates clustering performance, for both zero-shot and with prompting. 
By contrast, increasing Qwen's size from 0.6 billion to 4 billion parameters raises last-token accuracy from 30.7\,\% to 46.3\,\%, matching Llama’s performance. However, a further increase to 8 billion parameters yields no additional gains in any configuration. Thus, for Qwen, the benefits of scaling plateau before the 8 billion parameter mark, whereas Llama shows minor sensitivity to model size in this task.

\subsection{Detailed Performance and Standard Deviation of the Best Model}
\label{sec:appendix_detailed}

\begin{table}[h]
  \centering
  \small
  \begin{tabular}{lr}
    \hline
    \textbf{Dataset/Task} & \textbf{Llama\_last} \\
    \hline
    ArXivHierarchicalClusteringP2P     & 62.71 $\pm$  \phantom{0}2.34 \% \\
    ArXivHierarchicalClusteringS2S     & 63.00 $\pm$  \phantom{0}4.13 \% \\
    BigPatentClustering.v2             & 32.11 $\pm$  \phantom{0}3.44 \% \\
    BiorxivClusteringP2P.v2            & 42.35 $\pm$  \phantom{0}0.57 \% \\
    BiorxivClusteringS2S.v2            & 38.80 $\pm$  \phantom{0}0.56 \% \\
    BuiltBenchClusteringP2P            & 58.22 $\pm$ 12.50 \% \\
    BuiltBenchClusteringS2S            & 50.69 $\pm$ 10.74 \% \\
    ClusTREC-Covid                     & 84.89 $\pm$  \phantom{0}0.45 \% \\
    MasakhaNEWSClusteringP2P           & 64.15 $\pm$ 23.64 \% \\
    MasakhaNEWSClusteringS2S           & 59.09 $\pm$ 23.70 \% \\
    MedrxivClusteringP2P.v2            & 39.14 $\pm$  \phantom{0}0.78 \% \\
    MedrxivClusteringS2S.v2            & 37.21 $\pm$  \phantom{0}0.68 \% \\
    RedditClustering.v2                & 58.91 $\pm$  \phantom{0}0.85 \% \\
    RedditClusteringP2P.v2             & 59.72 $\pm$  \phantom{0}0.37 \% \\
    SIB200ClusteringS2S                & 54.67 $\pm$  \phantom{0}5.04 \% \\
    StackExchangeClustering.v2         & 65.67 $\pm$  \phantom{0}0.75 \% \\
    StackExchangeClusteringP2P.v2      & 43.73 $\pm$  \phantom{0}0.43 \% \\
    TwentyNewsgroupsClustering.v2      & 53.03 $\pm$  \phantom{0}1.50 \% \\
    WikiCitiesClustering               & 80.68 $\pm$  \phantom{0}0.00 \% \\
    WikipediaChemistryTopicsClustering & 67.47 $\pm$  \phantom{0}0.00 \% \\
    WikipediaSpecialtiesInChemistryCl. & 28.08 $\pm$  \phantom{0}0.00 \% \\
    \hline
    AVG & \textbf{54.49} \% \\
    \hline
  \end{tabular}
  \caption{Clustering accuracy of \texttt{Llama\_last} across multiple MTEB tasks. Results show the \texttt{v\_measure} and \texttt{v\_measure\_std} reported by running MTEB using mteb.get\_tasks(task\_types="Clustering", languages=["eng"]).}
  \label{tab:precise_values_std_deviation}
\end{table}

In Table \ref{tab:precise_values_std_deviation} we report per-dataset v-measure and standard deviation for all 21 English clustering tasks in MTEB v1.38.30. The ± values are not derived from repeated experimental runs, but from the variability induced by MTEB’s own evaluation procedure. Specifically, for clustering tasks MTEB fixes the dataset splits and embeddings, but the downstream clustering evaluation involves multiple runs (e.g., different initializations or folds), and the reported standard deviation quantifies this within-task variation. It should therefore be interpreted as a measure of how sensitive the evaluation metric is to these internal clustering runs, not as a statistical confidence interval. The benchmark suite spans diverse domains, including scientific abstracts (ArXiv, BioRxiv, MedRxiv), news (MasakhaNEWS), online forums (Reddit, StackExchange), and encyclopedic text (Wikipedia, WikiCities). It further distinguishes between task granularities, such as Sentence-to-Sentence (S2S) tasks, which use short text segments like titles or individual sentences, and Passage-to-Passage (P2P) tasks, which rely on longer text spans such as abstracts or full passages \cite{muennighoff2023mtebmassivetextembedding}. This coverage ensures that the results reflect robustness across heterogeneous text types. The macro-average of 54.49\,\% indicates stable overall performance, with the strongest results on well-structured corpora such as ClusTREC-Covid and WikiCities. When comparing tasks available in both configurations, P2P generally outperforms S2S, as observed for BioRxiv, BuiltBench, MasakhaNEWS, MedRxiv, and Reddit. However, this trend is not universal: ArXiv shows nearly identical performance for both settings, while StackExchange exhibits a strong advantage for S2S over P2P. These results indicate that the optimal task granularity is dataset-dependent, with P2P often providing richer contextual signals but S2S remaining competitive or superior when concise text segments are more informative. Different tasks reach their best performance after varying numbers of training iterations. Because evaluation was conducted jointly across all 21 tasks, checkpoints were selected based on overall performance rather than task-specific peaks. As a result, the chosen checkpoints may not be optimal for every individual dataset. All results were obtained with the official MTEB evaluation pipeline \cite{muennighoff2023mtebmassivetextembedding}, ensuring direct comparability with prior work.

\subsection{Analysis of Attention}
\label{sec:attention}

In this section, we have chosen randomly several additional example sentences for a visual attention analysis. Each plot shows the average attention received by every input token from the final token in the last transformer layer of Llama-3.2-1B. For the fine-tuned condition, we use the best-performing model. Bars indicate attention at three stages of training: prompt-only, after 100 fine-tuning steps, and after 20,000 steps. Across all examples, the pattern is consistent: short fine-tuning shifts attention from prompt tokens toward the actual sentence being encoded, improving over the prompt-only baseline, whereas prolonged fine-tuning results in a more uniform distribution. The sentences were chosen to be short for readability. In addition to examples aligned with \cite{Simple_Techniques}, we include samples from the EmotionClassification dataset \cite{muennighoff2023mtebmassivetextembedding} as well as other simple sentences.

\begin{figure*}[t]
  \centering
  \includegraphics[width=0.8\textwidth]{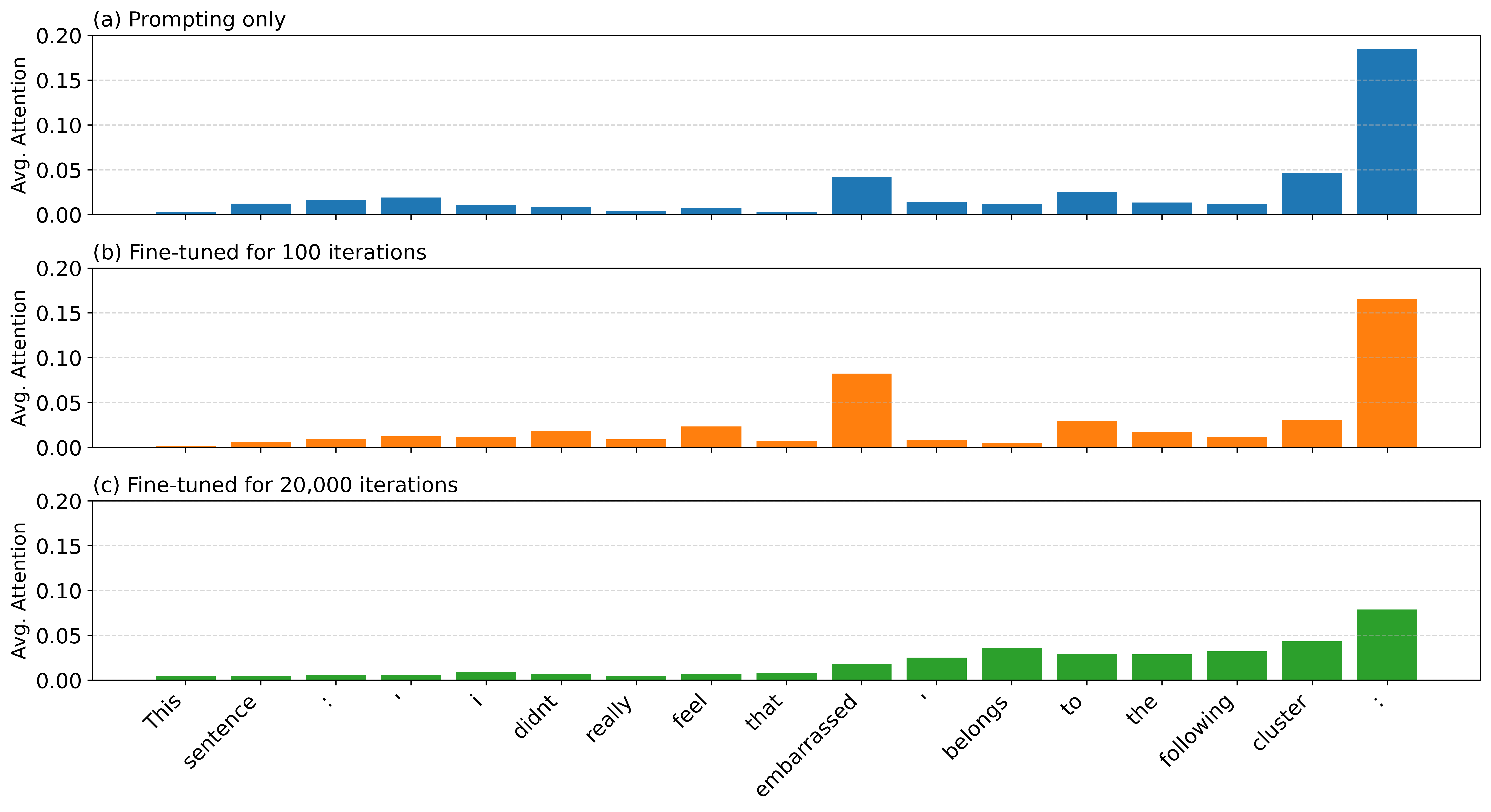}

  \medskip

  \includegraphics[width=0.8\textwidth]{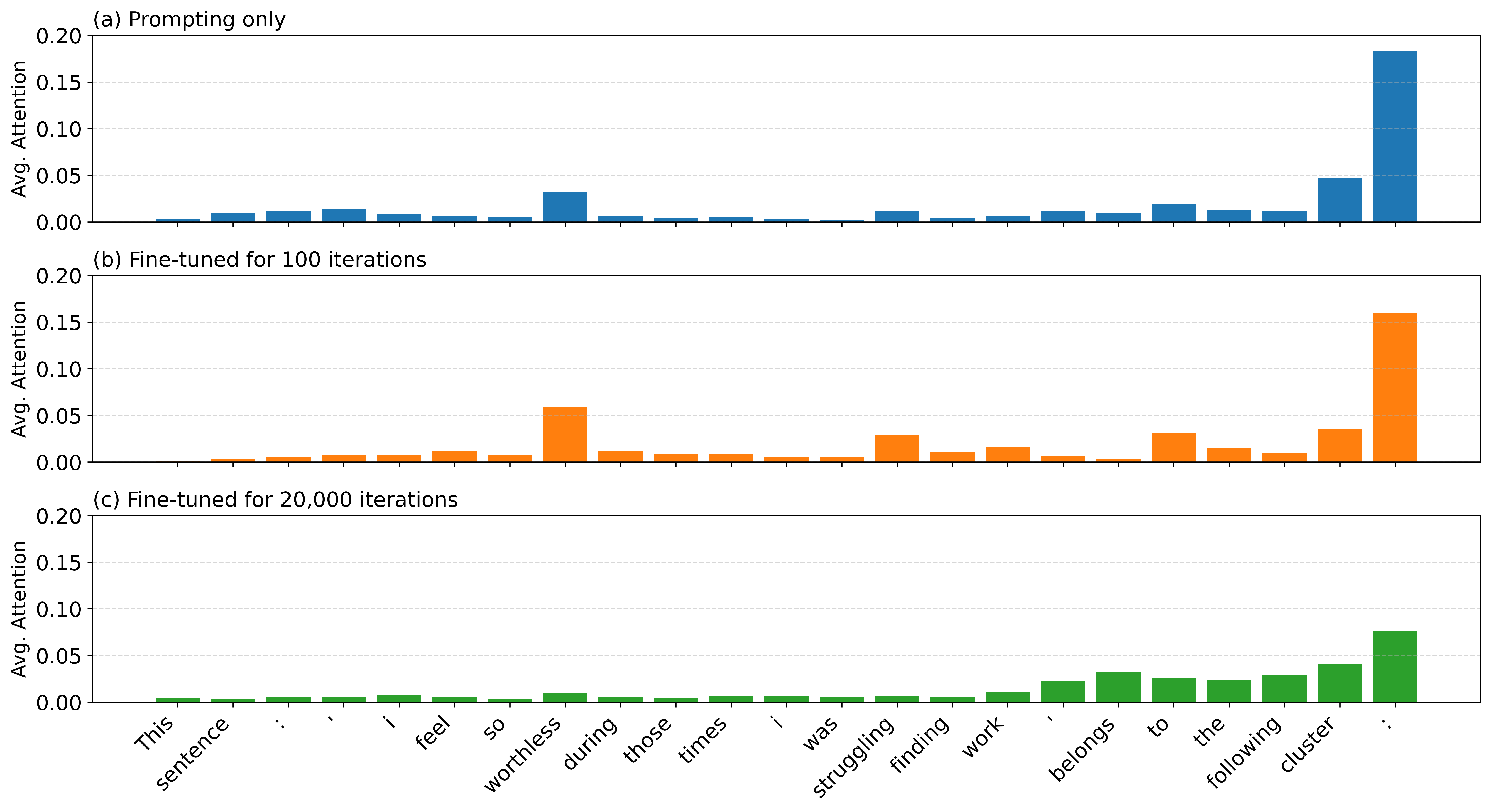}
  \caption{Average attention received by each input token from the final token in the last layer of Llama-3.2-1B with the CCW prompt. Shown for two EmotionClassification sentences \cite{muennighoff2023mtebmassivetextembedding}. Each plot reports three stages: prompt-only, after 100 fine-tuning steps, and after 20{,}000 steps.}
  \label{fig:attention_comparison-double_1}
\end{figure*}

\begin{figure*}[t]
  \centering
  \includegraphics[width=0.8\textwidth]{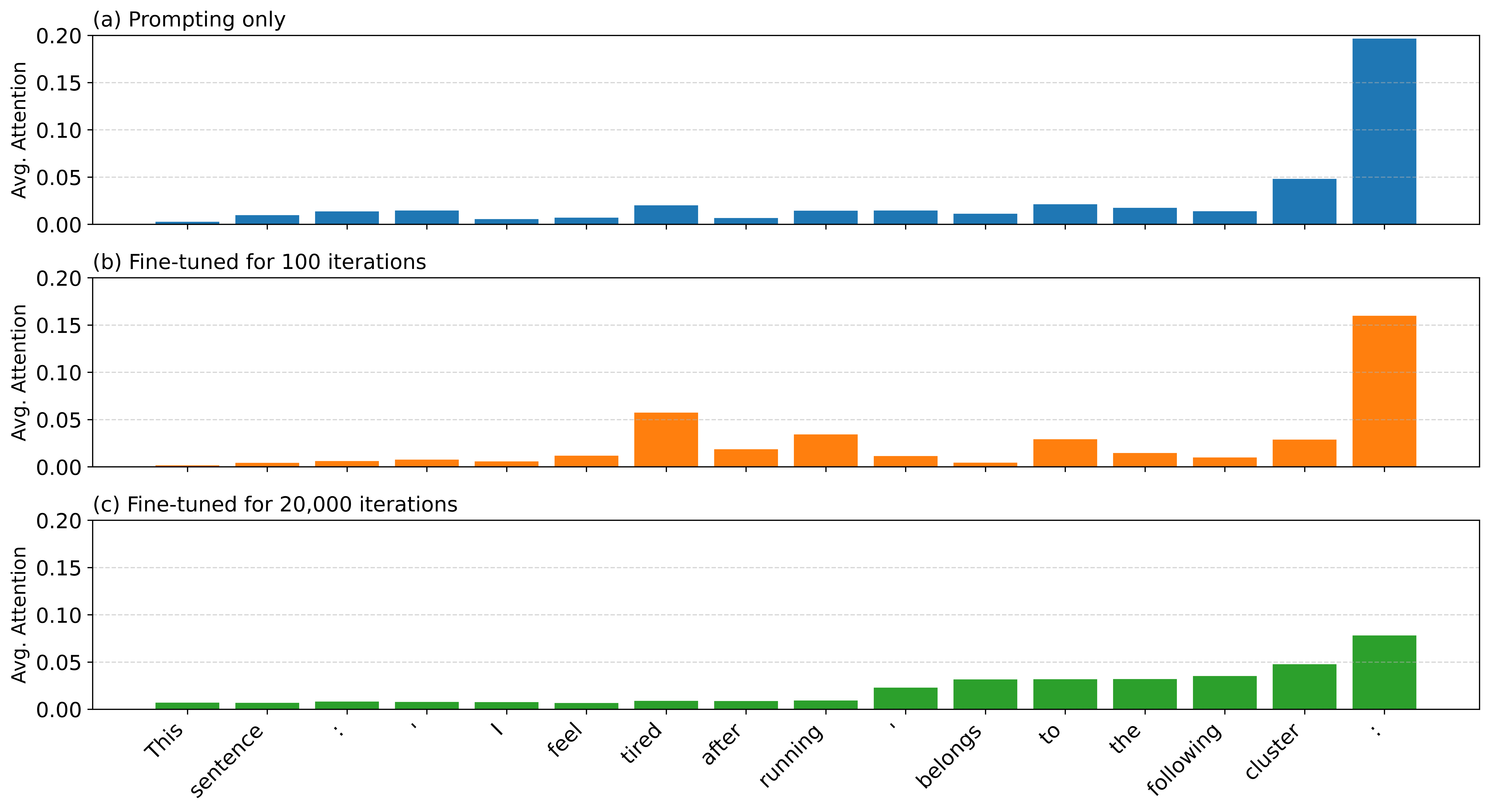}

  \medskip

  \includegraphics[width=0.8\textwidth]{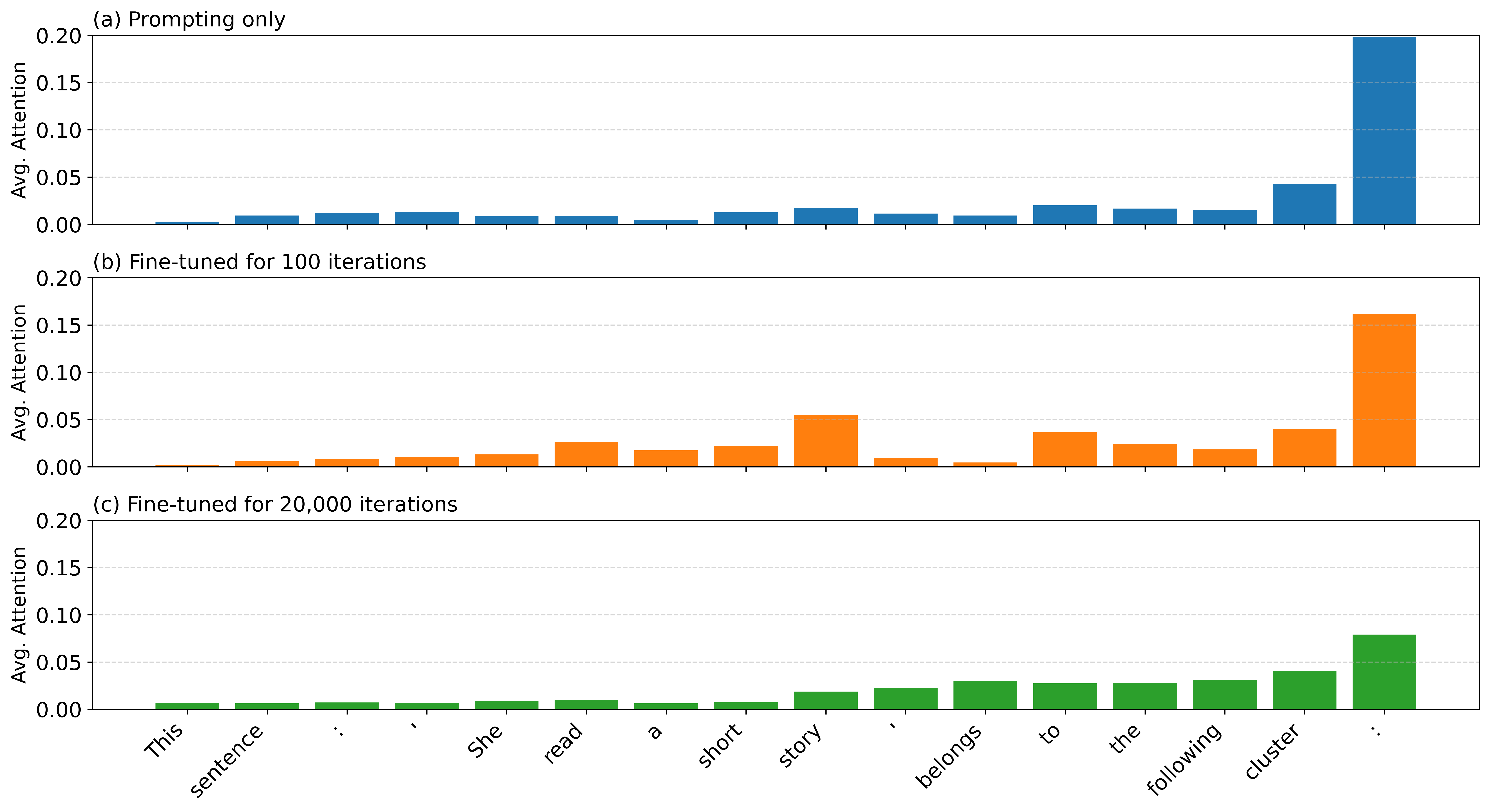}
  \caption{Average attention received by each input token from the final token in the last layer of Llama-3.2-1B with the CCW prompt. Shown for two random short sentences: “I feel tired after running” and “She read a short story.” Each plot reports three stages: prompt-only, after 100 fine-tuning steps, and after 20{,}000 steps.}
  \label{fig:attention_comparison-double_2}
\end{figure*}

Figures~\ref{fig:attention_comparison-double_1} and~\ref{fig:attention_comparison-double_2} 
show how fine-tuning changes attention patterns across four sentences. For the EmotionClassification 
examples, attention shifts from prompt tokens toward emotionally salient words such as 
\emph{embarrassed}, \emph{worthless}, and \emph{struggling}. In the shorter generic sentences, 
the model similarly learns to emphasize core content words like \emph{tired}, \emph{running}, 
\emph{read}, and \emph{story}. Extended training eventually redistributes attention more evenly 
while still retaining stronger focus on semantically central tokens.

Overall, the zero-shot setting already reveals that semantically relevant tokens tend to attract higher attention, indicating that the model encodes some notion of relevance without task-specific adaptation. Fine-tuning maintains this advantage while shifting focus away from prompt tokens toward the sentence to be embedded. The combination of prompt usage, which guides the model to store information in the final token, and fine-tuning, which encourages greater focus on semantic content, indicates that the model learns to produce more discriminative embeddings.

\end{document}